\documentclass[10pt,twocolumn,letterpaper]{article}

\usepackage{cvpr}
\usepackage{times}
\usepackage{epsfig}
\usepackage{graphicx}
\usepackage{amsmath}
\usepackage{amssymb}
\usepackage{comment}
\usepackage{threeparttable}  
\usepackage{url}     
\usepackage{algorithm}
\usepackage{algpseudocode} 
\usepackage{subfig}
\usepackage{caption}
\usepackage{multirow}
\usepackage{multicol}
\usepackage{booktabs}

\usepackage{epstopdf}
\AppendGraphicsExtensions{.tif}

\usepackage[pagebackref=true,breaklinks=true,letterpaper=true,colorlinks,bookmarks=false]{hyperref}

\cvprfinalcopy 

\def\cvprPaperID{****} 

\ifcvprfinal\fi
\begin{document}

\title{The Multi-Temporal Urban Development SpaceNet Dataset}


\author{Adam Van Etten$^{1}$, 
Daniel Hogan$^{1}$, 
Jesus Martinez-Manso$^{2}$,
Jacob Shermeyer$^{3}$,
Nicholas Weir$^{4, \dagger}$,
Ryan Lewis$^{4,\dagger}$\\
\\
$^{1}$ In-Q-Tel CosmiQ Works, [avanetten, dhogan]@iqt.org, 
$^{2}$ Planet, jesus@planet.com,\\
$^{3}$ Capella Space, jake.shermeyer@capellaspace.com,
$^{4}$ Amazon, [weirnich, rstlewis]@amazon.com \\
}

\maketitle

\begin{abstract}
Satellite imagery analytics have numerous human development and disaster response applications, particularly when time series methods are involved. For example, quantifying population statistics is fundamental to 67 of the 231 United Nations Sustainable Development Goals Indicators, but the World Bank estimates that over 100 countries currently lack effective Civil Registration systems. To help address this deficit and develop novel computer vision methods for time series data, we present the Multi-Temporal Urban Development SpaceNet (MUDS, also known as SpaceNet 7) dataset.  
This open source dataset consists of medium resolution (4.0m) satellite imagery mosaics, which includes $\approx24$ images (one per month) covering $>100$ unique geographies, and comprises $>40,000$ km$^2$ of imagery and exhaustive polygon labels of building footprints therein, totaling over 11M individual annotations.   Each building is assigned a unique identifier (\ie address), which permits tracking of individual objects over time.  Label fidelity exceeds image resolution; this ``omniscient labeling'' is a unique feature of the dataset, and enables surprisingly precise algorithmic models to be crafted.

We demonstrate methods to track building footprint construction (or demolition) over time, thereby directly assessing urbanization. Performance is measured with the newly developed SpaceNet Change and Object Tracking (SCOT) metric, which quantifies both object tracking as well as change detection. We demonstrate that despite the moderate resolution of the data, we are able to track individual building identifiers over time. This task has broad implications for disaster preparedness, the environment, infrastructure development, and epidemic prevention.

\end{abstract}

\section{Introduction} \label{sec:background}

Time series analysis of satellite imagery poses an interesting computer vision challenge, with many human development applications. We aim to advance this field through the release of a large dataset aimed 
at enabling new methods in this domain. Beyond its relevance for disaster response, disease preparedness, and environmental monitoring, time series analysis of satellite imagery poses unique technical challenges often unaddressed by existing methods. 

The MUDS dataset (also known as SpaceNet 7) consists of imagery and precise building footprint labels over dynamic areas for two dozen months, with each building assigned a unique identifier (see Section \ref{sec:data} for further details).  
In the algorithmic portion of this paper (Section \ref{sec:experiments}), we focus on tracking  building footprints to monitor construction and demolition in satellite imagery time series. We aim to identify all of the buildings in each image of the time series and assign identifiers to track the buildings over time.

Timely, high-fidelity foundational maps are critical to a great many domains. For example, high-resolution maps help identify communities at risk for natural and human-derived disasters. Furthermore, identifying new building construction in satellite imagery is an important factor in establishing population estimates in many areas (\eg \cite{Engstrom2019EstimatingSA}). Population estimates are also essential for assessing burden on infrastructure, from roads\cite{road_crashes} to medical facilities \cite{hospital_catchment}. 

The inclusion of unique building identifiers in the MUDS dataset enable potential improvements upon existing course population estimates.
Without unique identifiers building tracking is not possible; this means that over a given area one can only determine how many new buildings exist.  By tracking unique building identifiers one can determine which buildings changed (whose properties such as precise location, area, \etc can be correlated with features such as road access, distance to hospitals, etc.), thus providing a much more granular view into population growth.

\footnotetext[2]{This work was completed prior to Nicholas Weir and Ryan Lewis joining Amazon}
Several unusual features of satellite imagery (\textit{e.g.} small object size, high object density, dramatic image-to-image difference compared to frame-to-frame variation in video object tracking, different color band wavelengths and counts, limited texture information, drastic changes in shadows, and repeating patterns) are relevant to other tasks and data. For example, pathology slide images or other microscopy data present many of the same challenges \cite{vwtas}. Lessons learned from this dataset may therefore have broad-reaching relevance to the computer vision community.

\section{Related Work}\label{sec:novelty}

\begin{figure}
    \centering
    \includegraphics[width=1.00\linewidth]{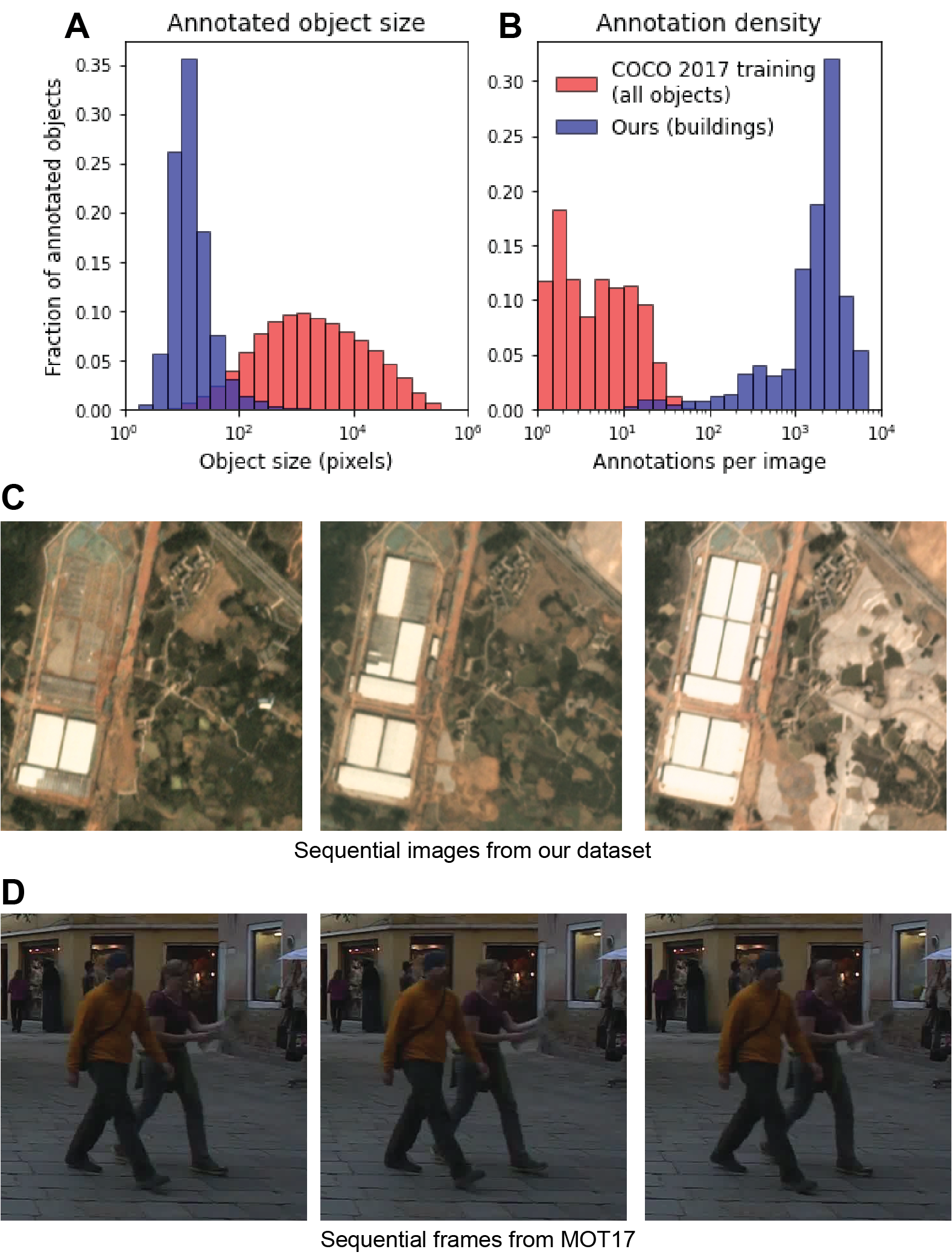}
    \caption{
    A comparison between our dataset and related datasets.
    \textbf{A.} Annotated objects are very small in this dataset. Plot represents normalized histograms of object size in pixels. Blue is our dataset, red represents all annotations in the COCO 2017 training dataset \cite{coco}. \textbf{B.} The density of annotations is very high in our dataset. In each $1024 \times 1024$ image, our dataset has between 10 and over 20,000 objects (mean: 4,600). By contrast, the COCO 2017 training dataset has at most 50 objects per image. \textbf{C.} Three sequential time points from one geography in our dataset, spanning 3 months of development. Compare to \textbf{D.}, which displays three sequential frames in the MOT17 video dataset \cite{MOT17}.}
    \label{fig:comparison}
    \vspace{-5pt}
\end{figure}

Past time series computer vision datasets and algorithmic advances have prepared the field to address many of the problems associated with satellite imagery analysis, allowing our dataset to explore additional computer vision problems. The challenge built around the VOT dataset \cite{VOT_TPAMI} saw impressive results for video object tracking (\textit{e.g.} \cite{Wang_2019_CVPR}), yet this dataset differs greatly from satellite imagery, with high frame rates and a single object per frame.  Other datasets such as MOT17 \cite{MOT17} or MOT20 \cite{mot20} have multiple targets of interest, but still have relatively few ($ < 20$) objects per frame. The Stanford Drone Dataset \cite{Robicquet2016LearningSE} appears similar at first glance, but has several fundamental differences that result in very different applications. That dataset contains overhead videos taken at multiple hertz  from a low elevation, and typically have $\approx20$ mobile objects (cars, people, buses, bicyclists, etc.) per frame. 
Because of 
the high frame rate of these datasets, frame-to-frame variation is minimal (see the MOT17 example in Figure \ref{fig:comparison}D). Furthermore, objects are larger and less abundant in these datasets than buildings are in satellite imagery. As a result, video competitions and models derived therein provide limited insight in how to manage imagery time series with substantial image-to-image variation and overly-dense instance annotations of target objects. Our data and research will address this gap.

To our knowledge, no existing dataset has offered a deep time series of satellite imagery. A number of previous works have studied building extraction from satellite imagery (\cite{sn123}, \cite{deepglobe}, \cite{sn4}, \cite{sn6}), yet these datasets were static. The closest comparison is the xView2 challenge and dataset \cite{xbd}, which examined building damage in satellite image pairs acquired before and after natural disasters (\ie only two timestamps) in $<20$ locations; however, this task fails to address the complexities and opportunities posed by analysis of deep time series data such as seasonal vegetation and lighting changes, or consistent object tracking on a global scale. Other competitions have explored time series data in the form of natural scene video, \textit{e.g.} object detection \cite{mot20} and segmentation \cite{Caelles_arXiv_2019} tasks. There are several meaningful dissimilarities between these challenges and the task described here. Firstly, frame-to-frame variation is very small in video datasets (see Figure \ref{fig:comparison}D). By contrast, the appearance of satellite images can change dramatically from month to month due to differences in weather, illumination, and seasonal effects on the ground, as shown in Figure \ref{fig:comparison}C. Other time series competitions have used non-imagery data spaced regularly over longer time intervals \cite{kaggle_google_traffic}, but none focused on computer vision tasks.

The size and density of target objects are very different in this dataset than past computer vision challenges. When comparing the size of annotated instances in the COCO dataset \cite{coco}, there's a clear difference in object size distributions (see Figure \ref{fig:comparison}A). These smaller objects intrinsically provide less information as they comprise fewer pixels, making their identification a more difficult task. Finally, the number of instances per image is markedly different in satellite imagery from the average natural scene dataset (see Section \ref{sec:data} and Figure \ref{fig:comparison}B).  Other data science competitions have explored datasets with similar object size and density, particularly in the microscopy domain \cite{recursion, BAH}; however, those competitions did not address time series applications.  Taken together, these differences highlight substantial novelty for this dataset. 

\section{Data}\label{sec:data}



\begin{figure}
    \centering
    \includegraphics[width=0.99\linewidth]{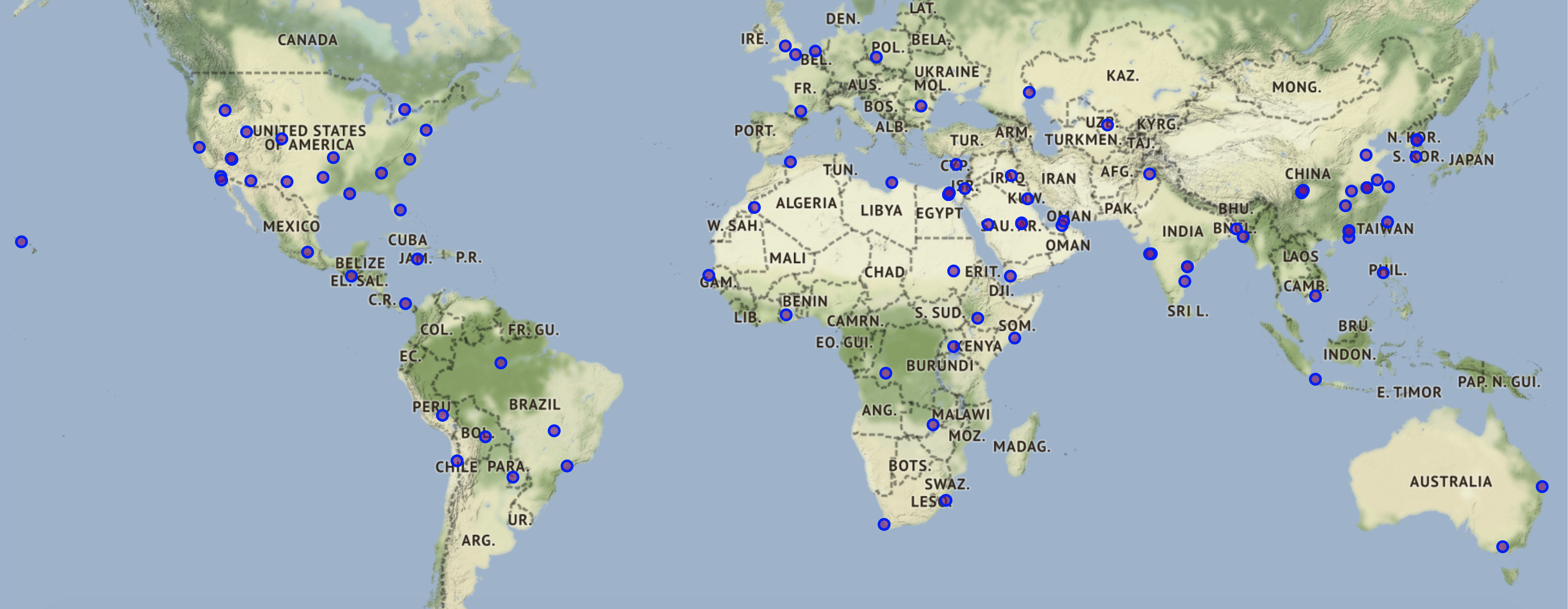}
    \vspace{-5pt}
    \caption{Location of MUDS data cubes.}
    \label{fig:map}
    \vspace{-7pt}
\end{figure}

The Multi-Temporal Urban Development SpaceNet (MUDS) dataset consists of 101 labelled sequences of satellite imagery collected by Planet Labs' Dove constellation 
between 2017 and 2020, coupled with building footprint labels for every image. The image sequences are sampled at the 101 distinct areas of interest (AOIs) across the globe, covering six continents (Figure \ref{fig:map}). These locations were selected to be both geographically diverse and display dramatic changes in urbanization across a two-year timespan.

The MUDS 
dataset is open sourced under a CC-BY-4.0 ShareAlike International license\footnote[3]{https://registry.opendata.aws/spacenet/} 
to encourage broad use. 
This dataset can potentially be useful for many other geospatial computer vision tasks: it can be easily fused or augmented with any other data layers that are available through web tile servers. The labels themselves can also be applied to any other remote sensing image tiles, such us high resolution optical or synthetic aperture radar. 




\subsection{Imagery}

Images are sourced from Planet's global monthly basemaps, an archive of on-nadir imagery containing visual RGB bands with a ground sample distance (GSD) (\ie pixel size) of $\approx 4$ meters. A basemap is a reduction of all individual satellite captures (also called scenes) into a spatial grid. These basemaps are created by mosaicing the best scenes over a calendar month, selected according to quality metrics like image sharpness and cloud coverage. Scenes are stack-ranked with best on top, and spatially harmonized to smoothen scene boundary discontinuities. 
Monthly basemaps are particularly well suited for the computer vision analysis of urban growth, since they are relatively cloud-free, homogeneous, and represented in a consistent spatio-temporal grid. The monthly cadence is also a good match to the typical timescale of urban developments. 

\begin{figure*}
    \centering
    \includegraphics[width=1.00\linewidth]{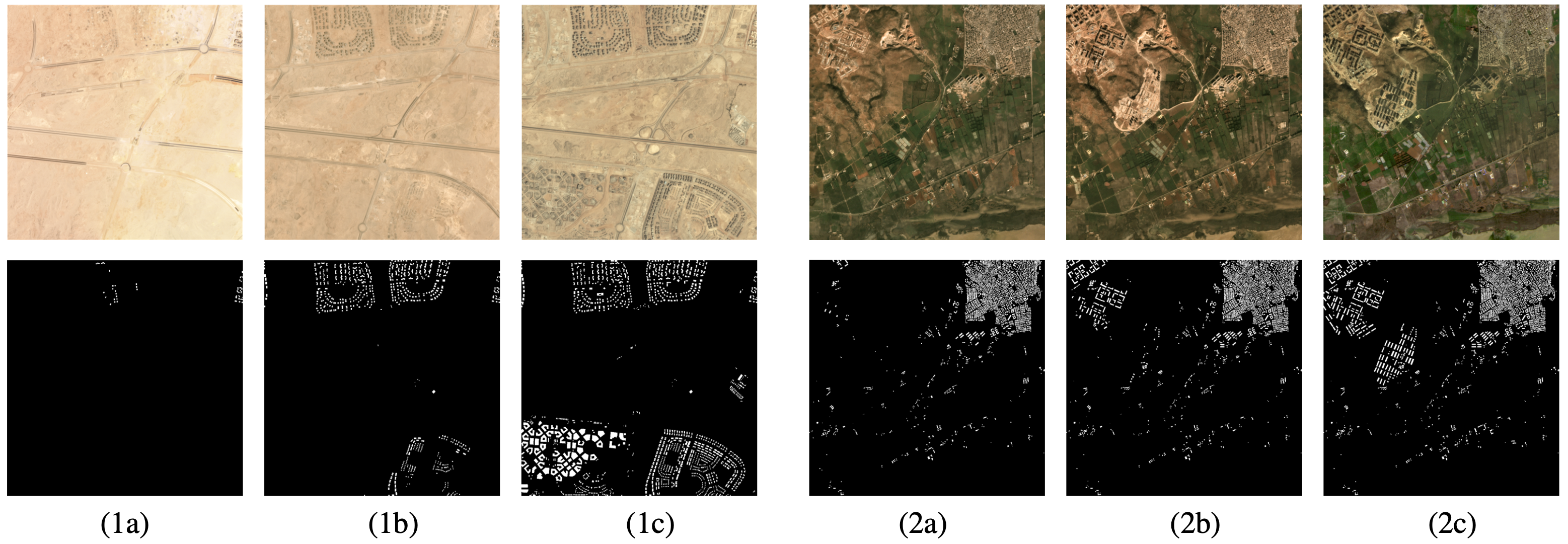}
    \vspace{-3pt}
    \caption{Time series of two data cubes.  Left column (\eg 1a) denotes the start of the times series, the middle column (\eg 1b) the approximate midpoint, and the right column (\eg 1c) shows the final image.  The top row displays imagery, while the bottom row illustrates the labeled building footprints.}
    \label{fig:grid}
    \vspace{-3pt}
\end{figure*}

The size of each image is $1024 \times 1024$ pixels, corresponding to $\approx18 \, \rm{km}^{2}$, and the total area of the images in the dataset is $41,250 \, \rm{km}^{2}$. See Table \ref{tab:datastats} or  \href{https://www.spacenet.ai}{spacenet.ai} for additional statistics. The time series contain imagery of $18-26$ months, depending on AOI (median of 24). This lengthy time span captures multiple seasons and atmospheric conditions, as well as the commencement and completion of multiple construction projects. See Figure \ref{fig:grid} for examples.  Images containing an excessive amount of clouds or haze were fully excluded from the dataset, thus causing minor temporal gaps in some of the time series.

\begin{figure}
\begin{tabular}{cc}
  \includegraphics[width=0.48\linewidth]{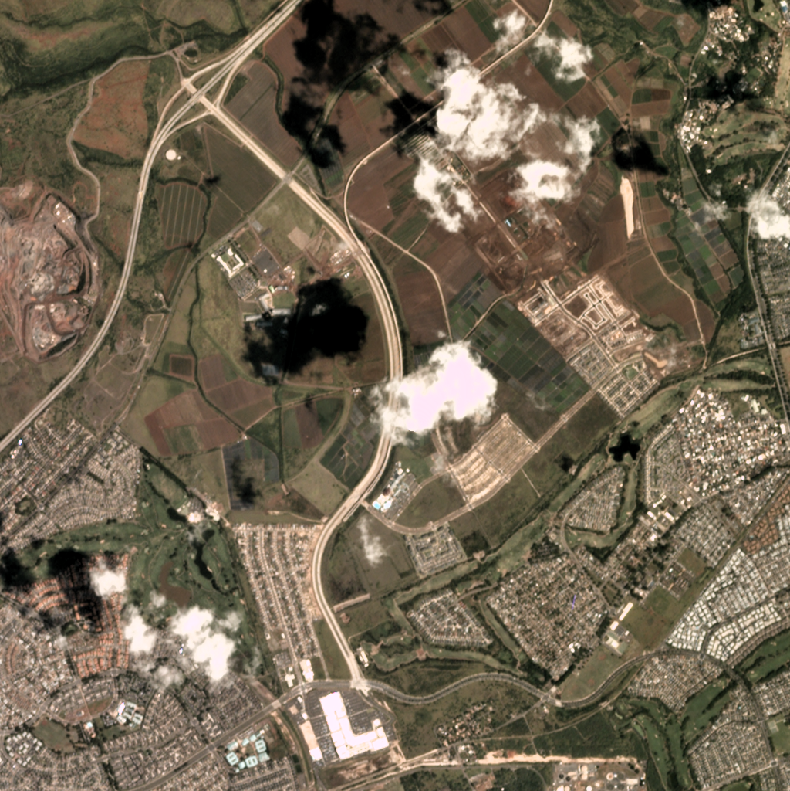} &   
  \includegraphics[width=0.48\linewidth]{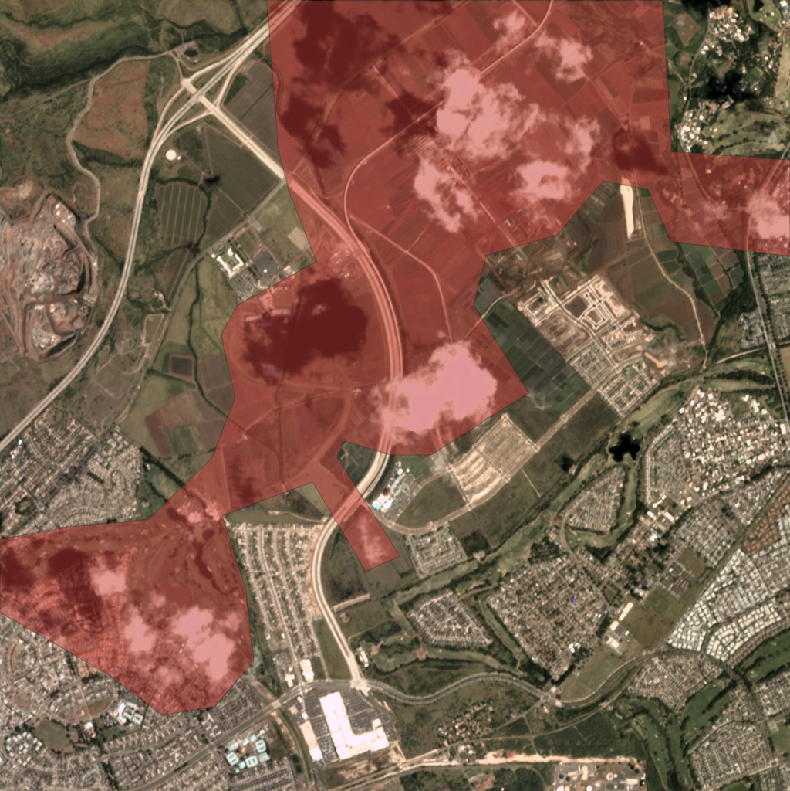} \\ 
  [-2pt] 
(a) Raw Image & (b) UDM overlaid \\  [4pt] 
 \includegraphics[width=0.48\linewidth]{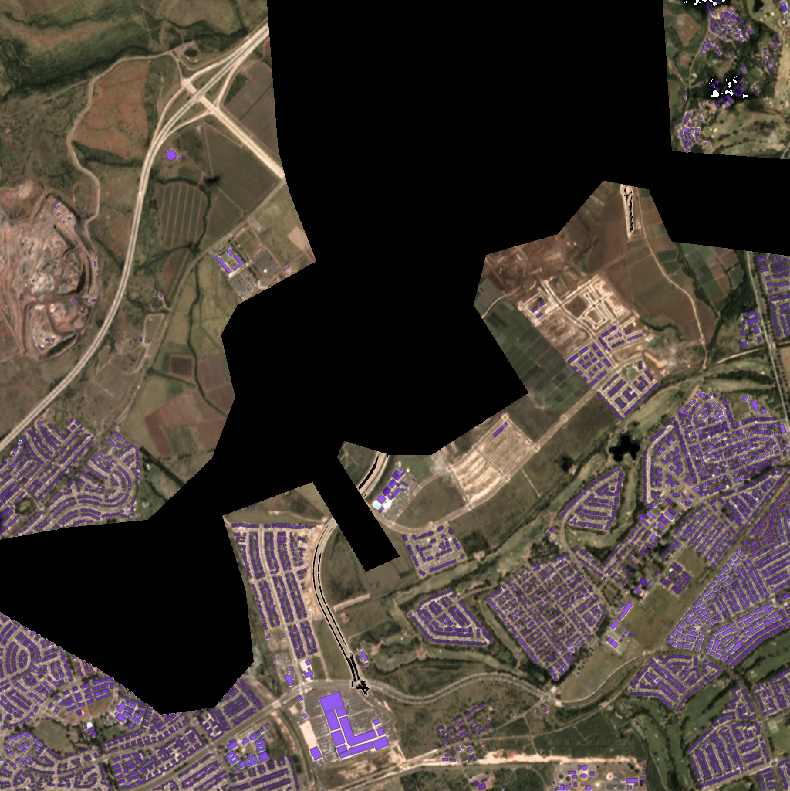} &  
 \includegraphics[width=0.48\linewidth]{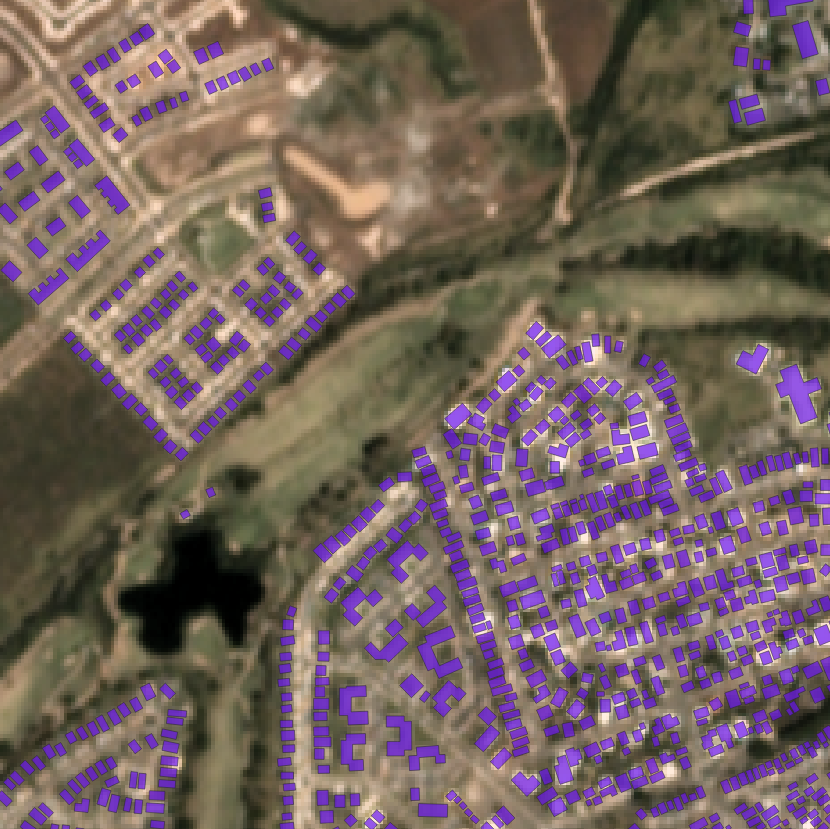} \\  
 [-2pt] 
(c) Masked image + labels & (d) Zoomed labels \\  [-0pt] 
\end{tabular}
\caption{Single image in a data cube.  (a) Image with cloud cover. (b) Image with UDM overlaid. (c) Masked image with building labels overlaid.  (d) Zoom showing the high fidelity of building labels.}
\label{fig:udm}
\vspace{-5pt}
\end{figure}

\subsection{Label Statistics}
 
Each image in the dataset is accompanied by two sets of manually created annotations. 
The first set of labels are building footprint polygons defining the outline of each building. 
Each building is assigned a unique identifier (\ie address) that persists throughout the time series. The second set of annotations are ``unusable data masks'' (UDMs) denoting areas of images that are obscured by clouds (see Figure \ref{fig:udm}) or that suffer from image geo-reference errors greater than 1 pixel. Geo-referencing is the process of mapping pixels in sensor space to geographic coordinates, performed via an empirical fitting procedure that is never exact. In rare cases, the scenes that compose the basemaps have spatial offsets of 5-10 meters. Accounting for such spatial displacements in the time series would make the modeling task significantly harder. Therefore, we decided to eliminate this complexity by including these regions in the UDM. 

Each image has between 10 and $\approx20,000$ building annotations, with a mean of $\approx4,600$
(the earliest timepoints in some geographies have very few buildings completed). 
This represents much higher label density than natural scene datasets like COCO \cite{coco} (Figure \ref{fig:comparison}B), or even overhead drone video datasets \cite{stanford_drone}. As the dataset comprises $\approx24$ time points at 101 geographic areas, the final dataset includes $>11$M annotations, 
representing $>500,000$ unique buildings. (Compare the training data quantities shown for other datasets in Table \ref{tab:datastats}.) The building areas vary between approximately 0.25 and 13,000 pixels (median building area of 193 m$^2$ or 12.1 pix$^2$),  markedly smaller than most labels in natural scene imagery datasets (Figure \ref{fig:comparison}A). 

Seasonal effects and weather (i.e. background variation) pervade our dataset given the low frame rate of $4\times10^{-7}$ Hz (Figure \ref{fig:comparison}C). This ``background'' change adds to the change detection task's difficulty. This frame-by-frame background variation is particularly unique and difficult to recreate via simulation or video re-sampling. 

\begin{table*}[t]
  \caption{Comparison of Selected Time Series Datasets}
  \vspace{-3pt}
  \label{tab:test_regs}
  \small
  \centering
   \begin{tabular}{lllllll}
    \hline
    & {\bf MUDS} & {\bf VOT-ST2020} & {\bf MOT20} & {\bf Stanford Drone} & {\bf DAVIS 2017} & {\bf YouTube-VOS}\\
    Property & & \cite{vot2020} & \cite{mot20} & \cite{Robicquet2016LearningSE} & \cite{davis2019} & \cite{youtubevos2018}\\
    \hline
    Scenes & 101 & 60 & 4 & 60 & 90 & 4,453\\
    Total Frames & 2,389 & 19,945 & 8,931 & 522,497 & 6,208 & $\sim$603,000 \\
    Unique Tracks & 538,073 & 60 & 2,332 & 10,300 & 216 & 7,755\\
    Total Labels & 11,079,262 & 19,945 & 1,652,040 & 10,616,256 & 13,543 & 197,272\\
    Median Frames/Scene & 24 & 257.5 & 1,544 & 11,008 & 70.5 & $\sim$135 (mean)\\
    Ground Sample Dist. & 4.0m & n/a & n/a & $\sim$2cm & n/a & n/a \\
    Frame Rate & 1/month & 30fps & 25fps & 30fps & 20fps & 30fps (6fps labels) \\
    Annotation & Polygon & Seg. Mask & BBox & BBox & Seg. Mask & Seg. Mask\\
    Objects & Buildings & Various & Pedestrians, & Pedestrians \& & Various & Various \\
    & & & etc. & Vehicles & & \\
  \end{tabular}
  \label{tab:datastats}
  \vspace{-5pt}
\end{table*}

\subsection{Labeling Procedure}

\begin{figure}
    \centering
    \includegraphics[width=0.99\linewidth]{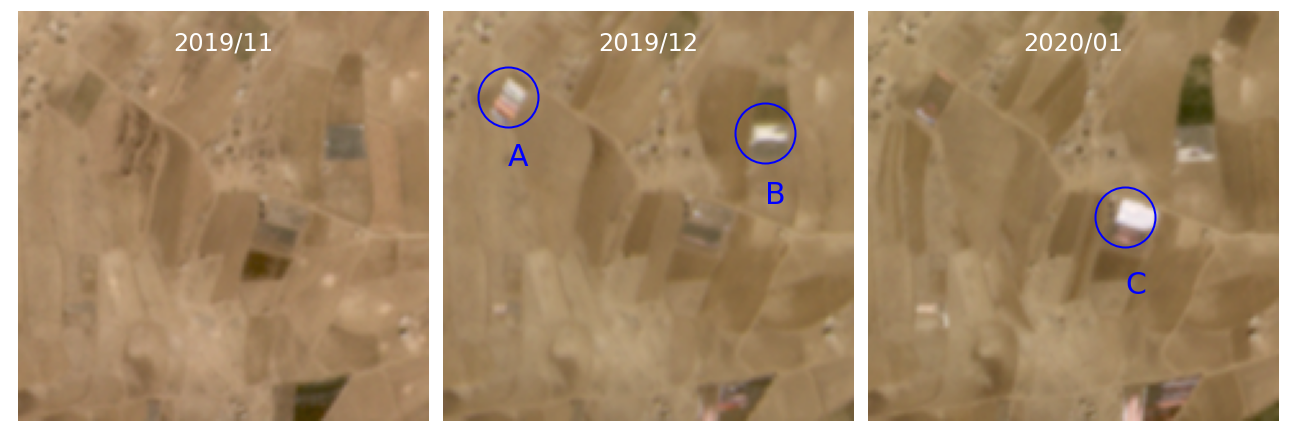}
    \vspace{-4pt}
    \caption{Example of how temporal context can help with object identification. If the middle image were to be labeled in isolation, objects A and B could be annotated as buildings. However, taking into account the adjacent images, these objects exist only for one month and therefore are unlikely to be buildings. Object C is also unlikely to be a building, just by group association.}
    \label{fig:ag_examples}
    \vspace{-7pt}
\end{figure}

We define buildings as static man-made structures where an individual could take shelter, with no minimum footprint size. 
The uniqueness of the dataset presents distinct labeling challenges.
First, small buildings can be under-resolved to the human eye in a given image, making it difficult to locate and discern from other non-building structures. Second, in locations undergoing  building construction, it can be difficult to determine what point in time the structure becomes a building per our definition. Third, variability in image quality, atmospheric conditions, shadows, and seasonal phenology can introduce additional confusion. Mitigating these complexities and minimizing label noise was of paramount importance, especially along the temporal dimension. Even though the dataset AOIs were selected to contain urban change, construction events are still highly imbalanced compared to the full spatio-temporal volume. Thus, temporal consistency was a fundamental area of focus in the labeling strategy. In cases of high uncertainty with a particular building candidate, annotators examined the full time series to gain temporal and contextual information of the precise location. For example, a shadow from a neighboring structure might be confused as a building, but this becomes evident when inspecting the full data cube. Temporal context can also help identify groups of objects. Some regions have structures that resemble buildings in a given image, but are highly variable in time. Objects that appear and disappear multiple times are unlikely to be buildings.
Once one type of such ephemeral structures is identified as a confusion source, all other similar structures are also excluded (Figure \ref{fig:ag_examples}). Labeling took 7 months by a team of 5; each data cube was annotated by one person, reviewed and corrected by another, with final validation by the team lead.

Annotators also used a privately-licensed high resolution imagery map to help discriminate uncertain cases. This high resolution map is useful to gain contextual information of the region and to guide the precise building outlines that are unclear from the dataset imagery alone. 
Once a building candidate was identified in the MUDS imagery, the high resolution map was used to confirm the building geometry. In other words, labels were not 
created on the high resolution imagery first. While the option of labeling on high resolution might seem attractive, it poses labeling risks such as capturing buildings that are not visible at all in the MUDS imagery. In addition, the high resolution map is static and composed of imagery acquired over a long range of dates, thus making it difficult to perform temporal comparisons between this map and the dataset imagery. 

The procedure to annotate each time series can be summarized as follows:
\begin{enumerate}
  \item Start with the first image in the series. Identify the location of all visible structures. 
  If the building location and outline are clear, draw a polygon around it. Otherwise, overlay a high resolution optical map to help confirm the presence of the building and draw the outline. Assign a unique integer identifier to each building. In addition, identify any regions in the image with impaired ground visibility or defects and add their polygons to the UDM layer of this image.
  \item Copy all the building labels onto the next image (not the UDM). Examine carefully all buildings in the new image, and edit the labels with any changes. Edits are only be made when there is significant confidence that a building appeared or disappeared. If a new building appeared, assign a new unique identifier.  Toggle through multiple images in the time series to ensure: (a) there is a true building change and (b) that it is applied to the correct time point. Also, create a UDM. 
  \item Repeat step 2 for the remaining time points.  
\end{enumerate}

\begin{figure}
    \centering
    \includegraphics[width=0.99\linewidth]{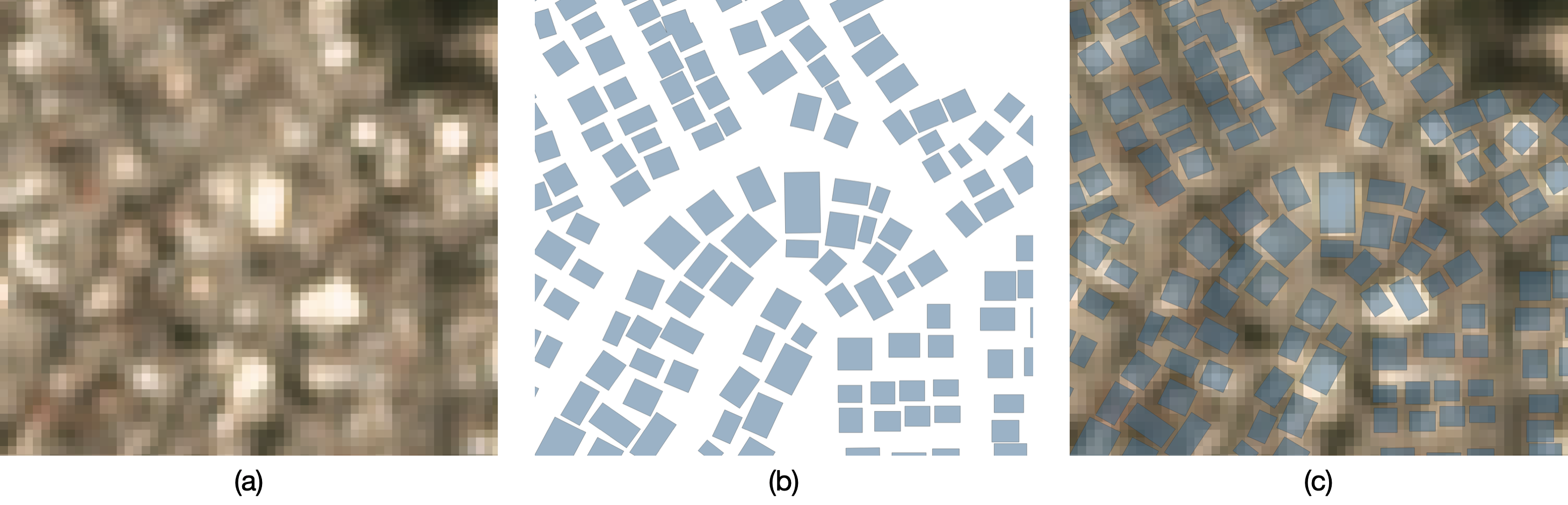}
    \caption{Zoom in of one particularly dense region illustrating the very high fidelity of labels. (a) Raw image. (b) Footprint polygon labels. (c) Footprints overlaid on imagery.}
    \label{fig:f33}
    \vspace{-5pt}
\end{figure}

This process attempts to enforce temporal consistency and reduce object confusion. While label noise is appreciable in small objects, the use of high resolution imagery to label results in labels of significantly higher fidelity that would be achievable from the Planet data alone, as illustrated in Figure \ref{fig:f33}.  This ``omniscient labeling'' is one of the key features of the MUDS dataset.  We will show in Section \ref{sec:experiments} that the baseline algorithm does a surprisingly good job of extracting high-resolution features from the medium-resolution imagery.  In effect, the labels are encoding information that is not visible to humans in the imagery, which the baseline algorithm is able to capitalize upon.

\section{Evaluation Metrics}\label{sec:metrics}

To evaluate model performance on a time series of identifier-tagged footprints such as MUDS, we introduce a new evaluation metric: the SpaceNet Change and Object Tracking (SCOT) metric \cite{SCOT}.  
As discussed later, existing metrics have a number of shortcomings that are addressed by SCOT.  The SCOT metric combines two terms: a tracking term and a change detection term. The tracking term evaluates how often a proposal correctly tracks the same buildings from month to month with consistent identifier numbers. In other words, it measures the model's ability to characterize what stays the same as time goes by. The change detection term evaluates how often a proposal correctly picks up on the construction of new buildings. In other words, it measures the model's ability to characterize what changes as time goes by.

For both terms, the calculation starts the same way: finding ``matches'' between ground truth building footprints and proposal building footprints for each month.  A pair of footprints (one ground truth and one proposal) are eligible to be matched if their intersection over union (IOU) exceeds 0.25, and no footprint may be matched more than once.  We select an IOU of 0.25 to mimic Equation 5 of ImageNet \cite{imagenet}), which sets IOU $< 0.5$ for small objects. A set of matches is chosen that maximizes the number of matches.  If there is more than one way to achieve that maximum, then as a tie-breaker the set with the largest sum of IOUs is used.  This is an example of the unbalanced linear assignment problem in combinatorics.

If model performance were being evaluated for a single image (instead of a time series), a customary next step might be calculating an F1 score, where matches are considered true positives ($tp$) and unmatched ground truth and proposal footprints are considered false negatives ($fn$) and false positives ($fp$) respectively.

\begin{equation}
F_1 = \frac{tp}{tp+\frac{1}{2}(fp+fn)}
\end{equation}

The tracking term and change detection term both generalize this to a time series, each in a different way.

The tracking term penalizes inconsistent identifiers across time steps.  A match is considered a ``mismatch'' if the ground truth footprint's identifier was most recently matched to a different proposal ID, or vice versa.  For the purpose of the tracking term, mismatches ($mm$) are not counted as true positives.  So each mismatch decreases the number of true positives by one.  This effectively divorces the ground truth footprint from its mismatched proposal footprint, creating an additional false negative and an additional false positive.  That amounts to the following transformations:
\begin{equation}
\begin{split}
tp&\rightarrow tp - mm \\
fp&\rightarrow fp + mm \\
fn&\rightarrow fn + mm
\end{split}
\end{equation}
Applying these to the F1 expression above gives the formula for the tracking term:
\begin{equation}
F_\text{track} = \frac{tp-mm}{tp+\frac{1}{2}(fp+fn)}
\end{equation}

The second term in the SCOT metric, the change detection term, incorporates only new footprints. That is, ground truth or proposal footprints with identifier numbers making their first chronological appearance.  Letting the subscript $new$ indicate the count of $tp$'s, $fp$'s, and $fn$'s that persist after dropping non-new footprints:
\begin{equation}
F_\text{change} = \frac{tp_\text{new}}{tp_\text{new}+\frac{1}{2}(fp_\text{new}+fn_\text{new})}
\end{equation}
One important property of this term is that a set of static proposals that do not vary from one month to another will receive a change detection term of 0, even for a time series with very little new construction.  (In the MUDS dataset, the construction of new buildings is by far the most common change; the metric could be generalized to accommodate building demolition or other changes by any of several straightforward generalizations.)

To compute the final score, the two terms are combined with a weighted harmonic mean:
\begin{equation}
F_\text{scot} = (1 + \beta^2) \frac{F_\text{change} \cdot F_\text{track}}{\beta^2 F_\text{change} + F_\text{track}}
\end{equation}
We use a value of $\beta=2$ to emphasize the part of the task (tracking) that has been less commonly explored in an overhead imagery context.
For a dataset like MUDS with multiple AOIs, the overall SCOT score is the arithmetic mean of the scores of the individual AOIs.

\begin{figure}
    \centering
    \begin{tabular}{cc}
    \includegraphics[width=0.47\linewidth]{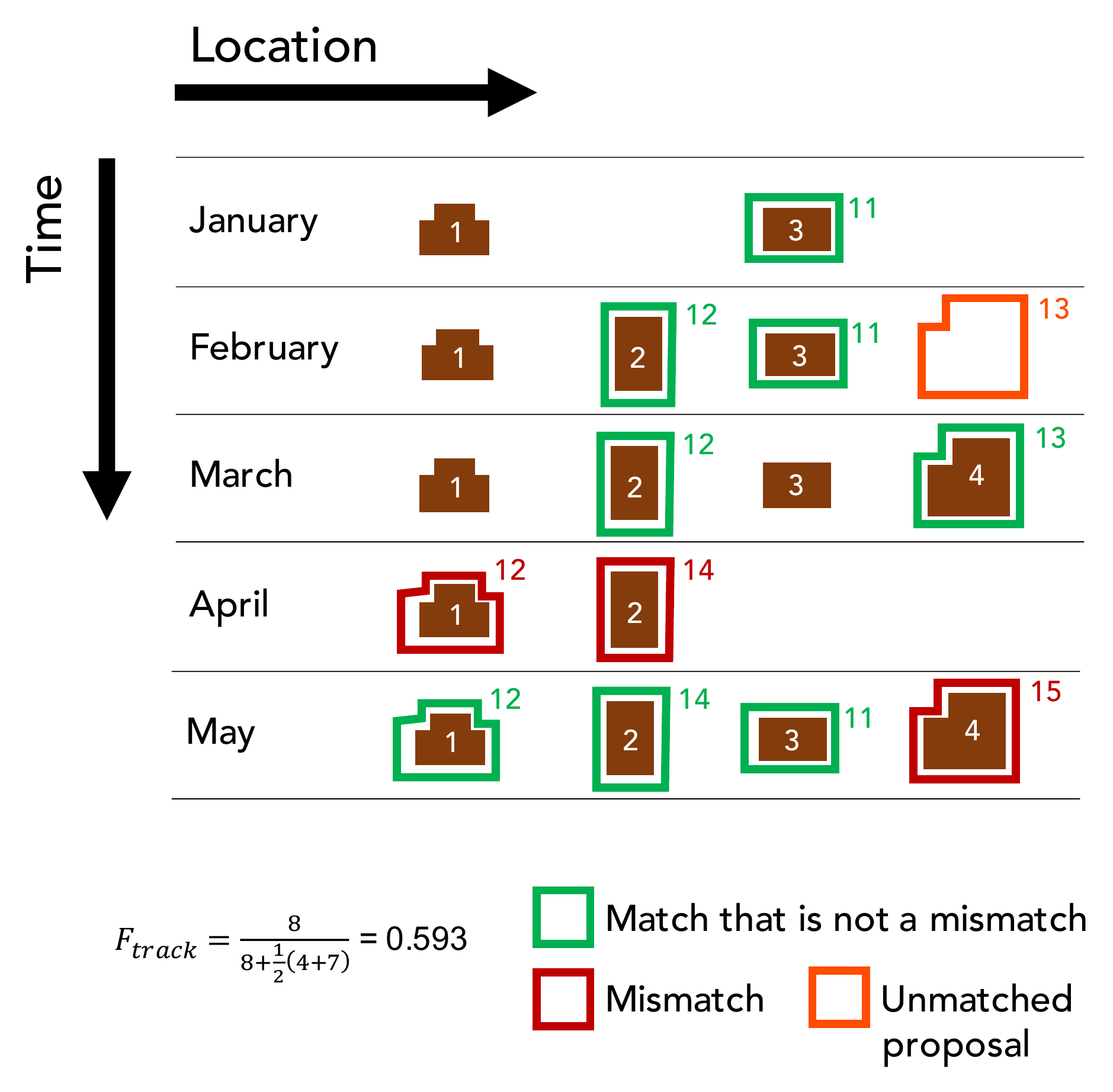} &   
    \includegraphics[width=0.47\linewidth]{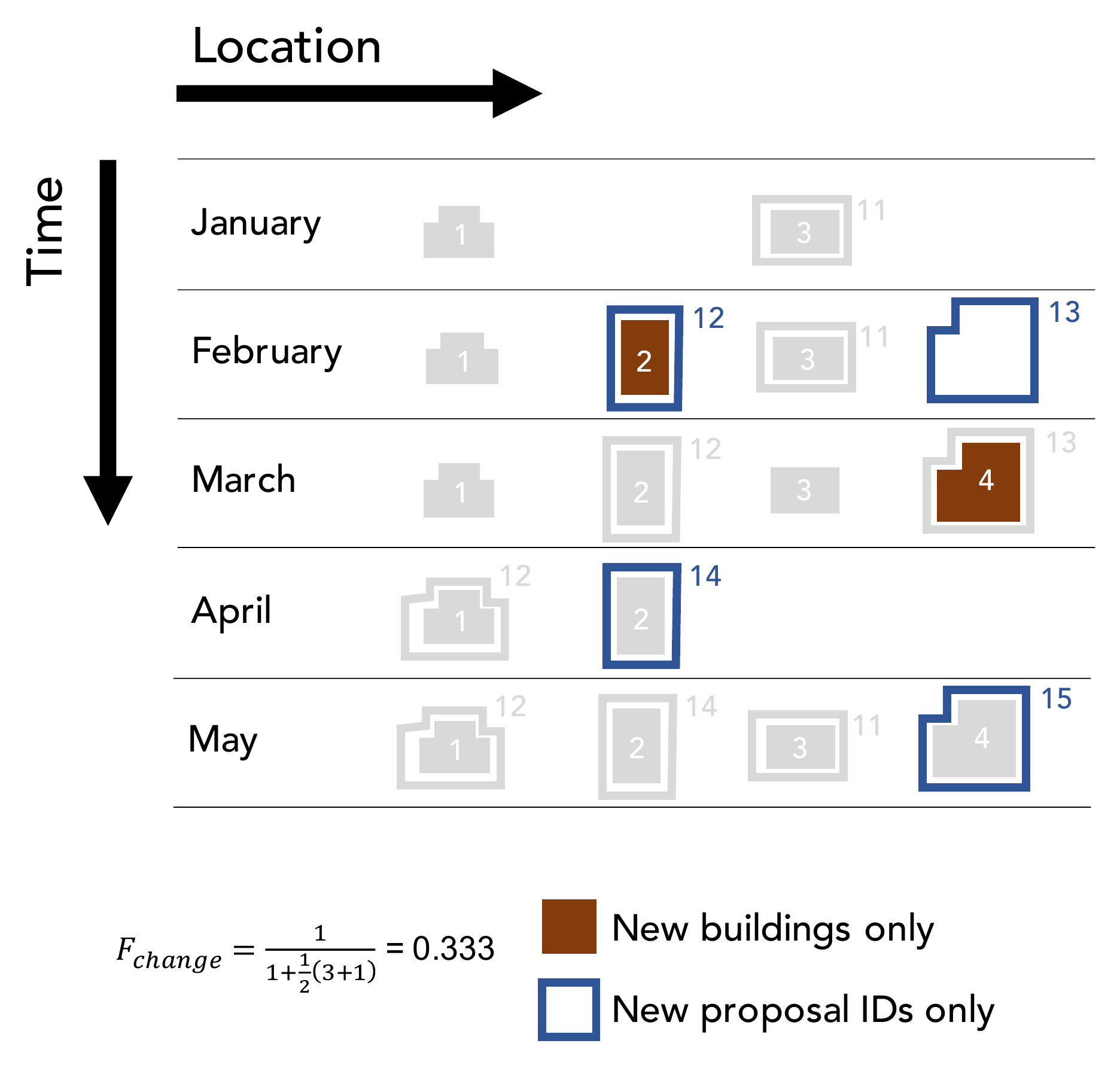} \\ 
    (a) Tracking Term & (b) Change Term \\ [-0pt] 
    \end{tabular}
    \caption{(a) Example of SCOT metric tracking term.  Solid brown polygons are ground truth building footprints, and outlines are proposal footprints.  Each footprint's corresponding identifier number is shown. (b) Example of SCOT metric change detection term, using the same set of ground truth and proposal footprints.  This term ignores all ground truth and proposal footprints with previously-seen identifiers, which are indicated in a faded-out gray color.}
    \label{fig:scot_meric}
    \vspace{-5pt}
\end{figure}

Figure \ref{fig:scot_meric}a 
is a cartoon example of calculating the tracking term on a row of four buildings imaged over five months (during which time two of the four are newly-constructed, and two are temporarily occluded by clouds).  
Figure \ref{fig:scot_meric}b 
illustrates the change detection term for the same case.

For geospatial work, the SCOT metric has a number of advantages over evaluation metrics developed for object tracking in video, such as the Multiple Object Tracking Accuracy (MOTA) metric \cite{MOTA}.  MOTA scores are mathematically unbounded, making them less intuitively interpretable for challenging low-score scenarios, and sometimes even yielding negative scores.  More critically, for scenes with only a small amount of new construction, it's possible to achieve a high MOTA score with a set of proposal footprints that shows no time-dependence whatsoever.  Since understanding time-dependence is usually a primary purpose of time series data, this is a serious drawback.  SCOT's change detection term prevents this.  In fact, many such approaches to ``gaming'' the SCOT metric by artificially increasing one term will decrease the other term, leaving no obvious alternative to intuitively-better model performance as a way to raise scores.



\section{Experiments}\label{sec:experiments}

For object tracking, one could in theory leverage the results of 
previous challenges (\eg MOT20 \cite{mot20}), yet the significant differences between MUDS and previous datasets such as high density and small object size (see Figure \ref{fig:comparison}) render previous approaches unsuitable.  For example, approaches such as TrackR-CNN \cite{mots} are untrainable as each instance requires a separate channel resulting in a memory explosion for images with many thousands of objects.  Other approaches such as Joint Detection and Embedding (JDE) \cite{realtime} are trainable; however inference results are ultimately incoherent due to the tiny object size and density overwhelming the YOLOv3 \cite{YOLOv3} detection grid.  Despite these challenges, the spatially static nature of our objects of interest somewhat simplifies tracking objects between each observation.   Consequently, this dataset should incentivize the development of new object tracking algorithms that can cope with a lack of resolution, spatial stasis, minimal size, and dense clustering of objects.

As a result of the challenges listed above, we choose to experiment with semantic segmentation based approaches to detect and track buildings over time.  These methods are adapted from prize winning approaches for the SpaceNet 4 and 6 Building Footprint Extraction Challenges \cite{MVOI, MSAW}. 
Our architecture comprises a U-Net \cite{u-net} with different encoders.  The first ``baseline'' approach uses a VGG16 \cite{vgg16} encoder and a custom loss function of $ \mathcal{L} = \mathcal{J} + 4 \cdot BCE$, 
where $\mathcal{J}$ is Jaccard distance and $BCE$ denotes binary cross entropy.  
The second approach uses a more advanced EfficientNet-B5 \cite{efficientnet} encoder with a loss of $ \mathcal{L} = \mathcal{F} + \mathcal{D}$ 
where $\mathcal{F}$ is Focal loss 
\cite{focal} and $\mathcal{D}$ is Dice loss.

To ensure robust testing statistics, we train the model on 60 data cubes, testing on the remaining 41 data cubes.  We train the segmentation models with an Adam optimizer on the 1424 images of the training set for 300 epochs and a learning rate of $10^{-4}$ (baseline) or 100 epochs and a learning rate of $2 \times 10^{-4}$ (EfficentNet).

\begin{figure}
    \centering
    \includegraphics[width=0.99\linewidth]{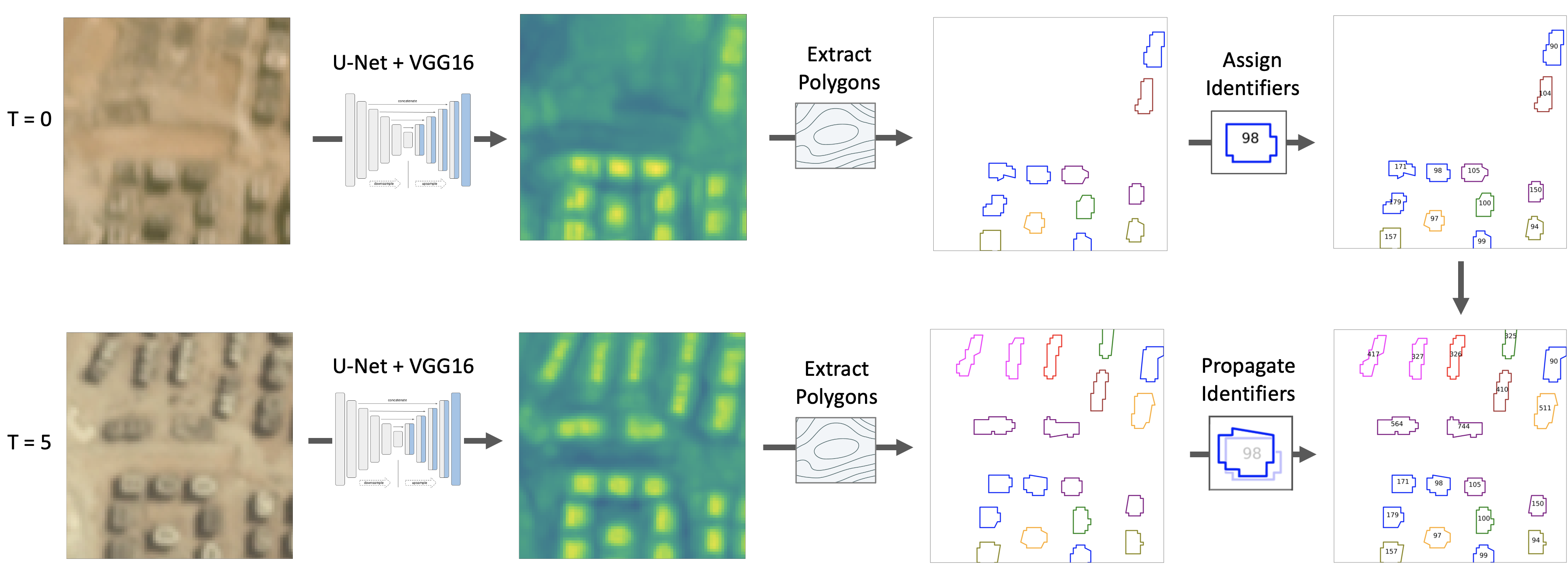}
    \caption{
    Baseline algorithm for building footprint extraction and identifier tracking showing evolution from T = 0 (top row) to T = 5 (bottom row).  The input image is fed into our segmentation model, yielding a building mask (second column).  This mask is refined into building footprints (third column), and unique identifiers are allocated (right column).}
    \label{fig:baseline_algo}
    \vspace{-2pt}
\end{figure}

At inference time binary building prediction masks are converted to instance segmentations of building footprints. Each footprint at $t=0$ is assigned a unique identifier.  For each subsequent time step building footprints polygons are compared to the positions of the previous time step.
Building identifier matching is achieved by an optimized matching of polygons with a minimum IOU overlap of 0.25.
Matched footprintes are assigned the same identifier as the previous timestep, while footprints without significant overlap with preceding geometries are assigned a new unique identifier. The baseline algorithm is illustrated in Figure \ref{fig:baseline_algo}; note that building identifiers are well matched between epochs. Performance 
is
summarized in Table \ref{tab:scores}.  For scoring we assess only buildings with area $\geq 4 \, \rm{px}^2$.

\begin{figure}
    \centering
    \includegraphics[width=0.99\linewidth]{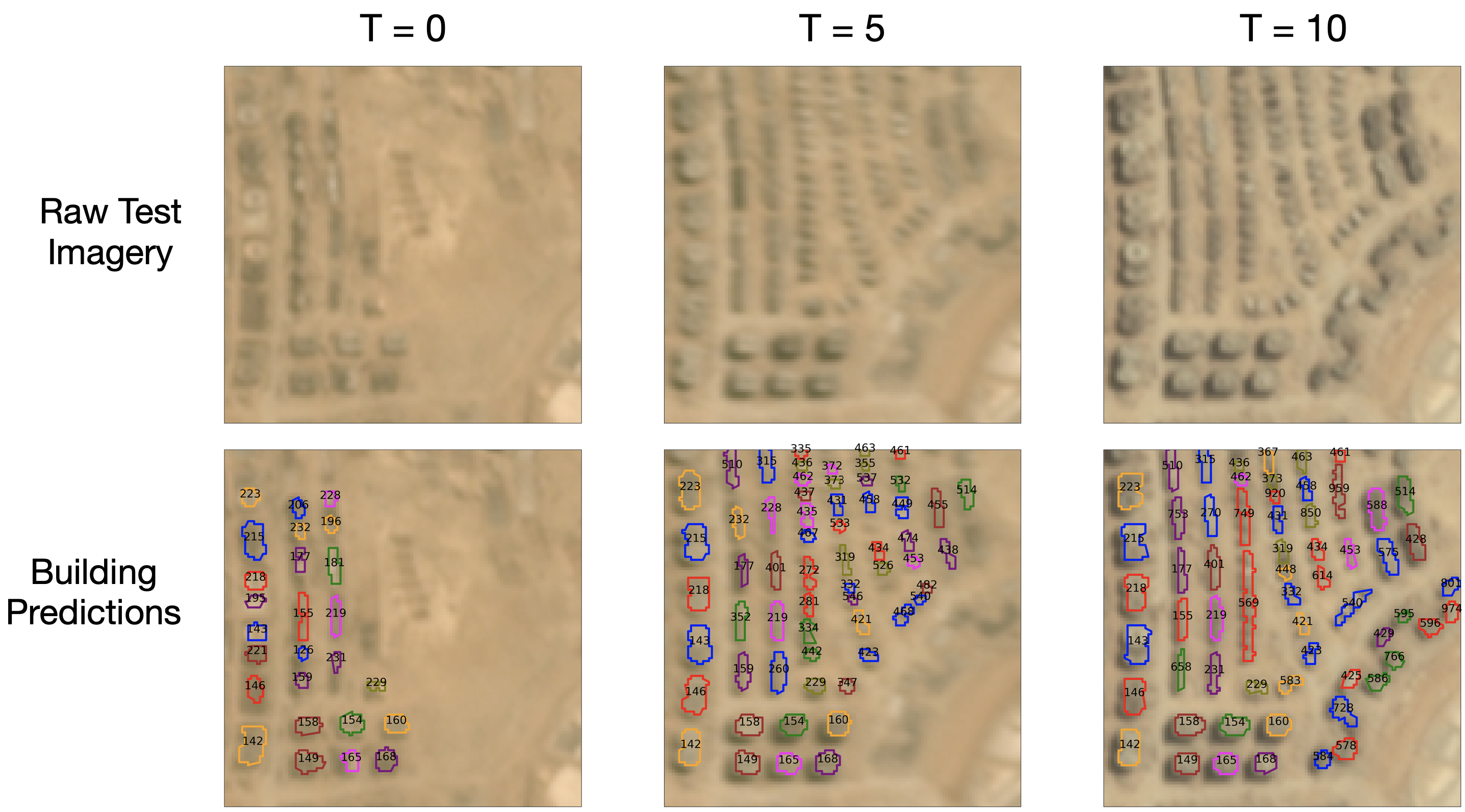}
    \caption{
    Example tracking performance of the baseline algorithm.  Note that larger, well-separated buildings are tracked well between epochs, while denser regions are more challenging for tracking.}
    \label{fig:f99}
    \vspace{-4pt}
\end{figure}

\begin{table}[]
  \caption{Building Tracking Performance}
  \label{tab:test_regs}
  \small
  \centering
   \begin{tabular}{lcc}
    \hline
    \multirow{2}{*}{\textbf{Metric}} &
  \multicolumn{2}{c}{\textbf{Approach}} \\ 
  \cline{2-3} &
  \multicolumn{1}{c}{\textbf{VGG-16}} &
  \multicolumn{1}{c}{\textbf{EfficentNet}} \\ 
    F1 (IOU $\geq 0.25$) &  $0.45 \pm 0.13$  &  $0.42 \pm 0.12$ \\
    Tracking Score &  $0.40 \pm 0.10$ &  $0.39 \pm 0.10$ \\
    Change Score &  $0.06 \pm 0.05$ &  $0.07 \pm 0.05$ \\
    SCOT &  $0.17 \pm 0.10$ &  $0.18 \pm 0.09$ \\
  \end{tabular}
  \vspace{-5pt}
  \label{tab:scores}
\end{table}

\begin{figure}
    \centering
    \includegraphics[width=0.99\linewidth]{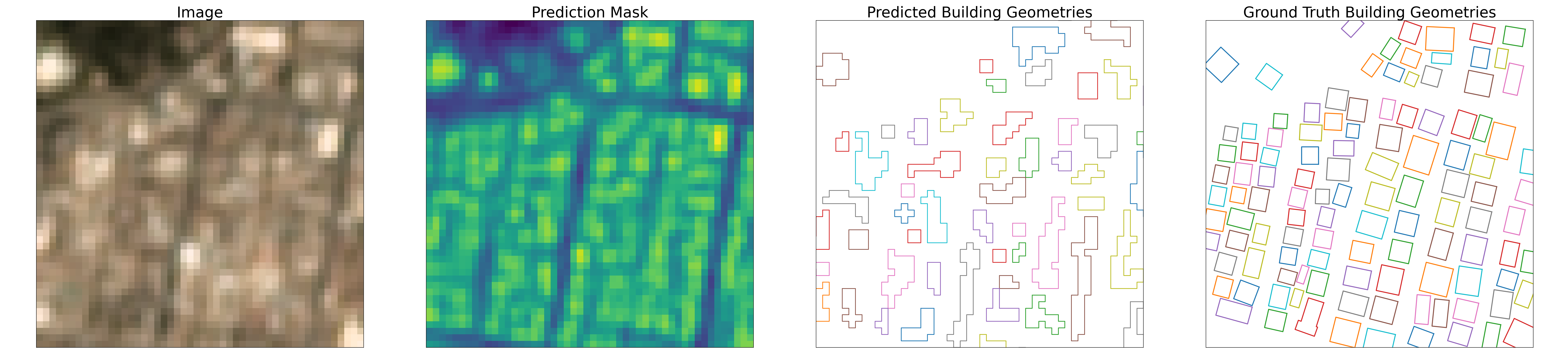}
    \caption{
    Prediction in a difficult, crowded region.  Despite the inherent difficulties in separating nearby buildings at medium resolution, for this image F1 = 0.40.}
    \label{fig:inf_zoom}
    \vspace{-6pt}
\end{figure}

Localizing and tracking buildings in medium resolution ($\approx4$m) imagery is quite challenging, but surprisingly achievable in our experiments.  For well separated buildings, building localization and tracking performs fairly well; for example in Figure \ref{fig:f99}) we find a localization F1 score of 0.55, and a SCOT score of 0.31.  For dense regions, building tracking is far more difficult; in Figure \ref{fig:inf_zoom} we still see decent performance in building localization (F1 = 0.40), yet building tracking and change detection is very challenging (SCOT = 0.07) since inter-epoch footprints overlap poorly.  The change term of SCOT is particularly challenging, as correctly identifying the origin epoch of each building is non-trivial, and spurious proposals are also penalized.

In an attempt to raise the scores of Table \ref{tab:scores}, we also endeavor to incorporate the time dimension into training.  As previously mentioned, existing approaches transfer poorly to this dataset, so we attempt a simple approach of stacking multiple images at training time.
For each date we train on the imagery for that date plus the four chronologically adjacent future observations [$t=0$, $t+1$, $t+2$, $t+3$, $t+4$] for five total dates of imagery.  When the number of remaining observations in the time series becomes less than five, we repeatedly append the final image for each area of interest. We find no improvement with this approach (SCOT $=0.17 \pm 0.08$).

We also note no significant difference in scores between the VGG-16 and EfficentNet architectures (Table \ref{tab:scores}), implying that older architectures are essentially as adept as state-of-the-art architectures when it comes to extracting information from the small objects in this dataset.

While not fully explored here, we also anticipate that researchers may improve upon the baseline using models specifically intended for time series analysis (\textit{e.g.} Recurrent Neural Networks (RNNs) \cite{RNN} and Long-Short Term Memory networks (LSTMs) \cite{LSTM}. In addition, numerous ``classical'' geospatial time series methods exist (\textit{e.g.} \cite{ZHU_2017}) which researchers may find valuable to incorporate into their analysis pipelines as well.

\section{Discussion}

Intriguingly, the score of $F1 = 0.45$ for our baseline mode parallels previous results observed in overhead imagery.  \cite{sr_earthvision} studied object detection performance in xView \cite{xview} satellite imagery for various resolutions and five different object classes.  These authors used the YOLT \cite{simrdwn} object detection framework, which uses a custom network based on the Googlenet \cite{googlenet} architecture.  The mean extent of the objects in this paper was 5.3 meters; at a resolution of 1.2 meters objects have an average extent of 4.4 pixels.  

The average building area for the MUDS dataset is 332 m$^2$, implying an extent of 18.2 m for a square object.  For a 4 meter resolution, this gives an average extent of 4.5 pixels, comparable to the 4.4 pixel extent of xView.  The observed MUDS F1 score of 0.45 is within error bars of the results of the xView results, see Table \ref{tab:comp}.  Of particular note is that while the F1 scores and object pixel sizes of Table \ref{tab:comp} are comparable, the datasets stem from vastly different sensors, and the techniques are wildly different as well (a Googlenet-based object detection architecture versus a VGG16-based segmentation architecture).  Apparently, object detection performance holds across sensors and algorithms as long as object pixel sizes are comparable.


\begin{table}[h]
  \caption{F1 Performance Across Datasets}
  \vspace{-3pt}
  \label{tab:test_regs}
  \small
  \centering
   \begin{tabular}{llll}
    \hline
    {\bf Dataset} & {\bf GSD (m)} & {\bf Object Size (pix)}  & {\bf F1} \\
    \hline
    xView & 1.2 & 4.4 & $0.41\pm0.03$ \\
    MUDS & 4.0 & 4.5 & $0.45\pm0.13$ \\
  \end{tabular}
  \label{tab:comp}
  \vspace{-5pt}
\end{table}

\section{Conclusions}

The Multi-temporal Urban Development SpaceNet (MUDS, also known as SpaceNet 7) dataset is a newly developed corpus of imagery and precise labels designed for tracking building footprints and unique identifiers.  The dataset covers over 100 locations across 6 continents, with a deep temporal stack of 24 monthly images and over 11,000,000 labeled objects. The significant scene-to-scene variation of the monthly images poses a challenge for computer vision algorithms, but also raises the prospect of developing algorithms that are robust to seasonal change and atmospheric conditions. One of the key characteristics of the MUDS dataset is exhaustive ``omniscient labeling'' with labels precision far exceeding the base imagery resolution of 4 meters.  Such dense labels present significant challenges in crowded urban environments, though we demonstrate surprisingly good building extraction, tracking, and change detection performance with our baseline algorithm.  Intriguingly, our object detection performance of $F1 = 0.45$ for objects averaging 4-5 pixels in extent is consistent with previous object detection studies, even though these studies used far different algorithmic techniques and datasets.  There are numerous avenues of research beyond the scope of this paper that we hope the community will tackle with this dataset: the efficacy of super-resolution, adapting video time-series techniques to the unique features of MUDS, experimenting with RNNs, Siamese networks, LSTMs, etc.  Furthermore, the dataset has the potential to aid a number of humanitarian efforts connected with population dynamics and UN sustainable development goals.

{\small
\bibliographystyle{ieee_fullname}
\bibliography{bib}
}

\end{document}